\documentclass{article}

    \PassOptionsToPackage{numbers, compress}{natbib}

\usepackage[preprint]{neurips_data_2023}

\usepackage[utf8]{inputenc} 
\usepackage[T1]{fontenc}    

\usepackage{url}            
\usepackage{booktabs}       
\usepackage{amsfonts}       
\usepackage{amsmath}       
\usepackage{nicefrac}       
\usepackage{microtype}      
\usepackage{xcolor}         
\usepackage{xspace}
\usepackage{bbm}
\usepackage{bm}
\usepackage{booktabs}
\usepackage{multirow}
\usepackage{subcaption}
\usepackage{enumitem}
\usepackage{marvosym}
\usepackage{tikz}
\usepackage{wrapfig}
\usepackage{forest}
\usepackage{subcaption}

\usepackage{hyperref}
\definecolor{mydarkblue}{rgb}{0,0.08,0.55}
\hypersetup{  
    colorlinks=true,
    linkcolor=mydarkblue,
    citecolor=mydarkblue,
    filecolor=mydarkblue,
    urlcolor=mydarkblue
}

\newcommand{\ind}{\mathbbm{1}}
\newcommand{\corr}{\text{corr}}

\def\eqref#1{equation~\ref{#1}}

\def\1{\bm{1}}

\def\rmX{{\mathbf{X}}}

\def\vx{{\bm{x}}}

\DeclareMathAlphabet{\mathsfit}{\encodingdefault}{\sfdefault}{m}{sl}
\SetMathAlphabet{\mathsfit}{bold}{\encodingdefault}{\sfdefault}{bx}{n}

\def\gC{{\mathcal{C}}}

\def\gS{{\mathcal{S}}}

\newcommand{\ptrain}{p_{\rm{train}}}
\newcommand{\ptest}{p_{\rm{test}}}

\newcommand{\R}{\mathbb{R}}

\usepackage{float}
\usepackage{pifont}
\newcommand{\xmark}{\textcolor{red}{\textbf{\ding{55}}}}
\newcommand{\cmark}{\textcolor{green}{\textbf{\ding{51}}}}

\definecolor{mydarkgreen}{RGB}{0, 100, 0}

\newcommand{\benchmark}{Spawrious\xspace}

\usepackage[many]{tcolorbox}    
\newtcolorbox{challenge}[1]{%
    tikznode boxed title,
    enhanced,
    boxrule=0.6mm,
    arc=1mm,
    interior style={white},
    attach boxed title to top center= {yshift=-\tcboxedtitleheight/2},
    fonttitle=\bfseries,
    colbacktitle=white,coltitle=black,
    boxed title style={size=normal,colframe=white,boxrule=0pt},
    title={#1}}
    
\usepackage[capitalize,noabbrev]{cleveref}

\newcommand{\red}[1]{\textbf{\textcolor{red}{#1}}}
\newcommand{\blue}[1]{\textbf{\textcolor{blue}{#1}}}
\newcommand{\green}[1]{\textbf{\textcolor{mydarkgreen}{#1}}}
\newcommand{\purple}[1]{\textbf{\textcolor{purple}{#1}}}
\newcommand{\orange}[1]{\textbf{\textcolor{orange}{#1}}}
\newcommand{\cyan}[1]{\textbf{\textcolor{cyan}{#1}}}

\newcommand{\domainbed}{\texttt{DomainBed}\xspace}

\title{Spawrious: A Benchmark for Fine Control \\of Spurious Correlation Biases}

\author{%
  Aengus Lynch\thanks{Equal contribution, alphabetical order.}\,\,, 
  Gbètondji J-S Dovonon\footnotemark[1]\,\, 
  Jean Kaddour\footnotemark[1]\,\,, 
  Ricardo Silva \\
  University College London \\
    \url{https://github.com/aengusl/spawrious}
}

\begin{document}

\maketitle

\begin{abstract}
The problem of spurious correlations (SCs) arises when a classifier relies on non-predictive features that happen to be correlated with the labels in the training data. For example, a classifier may misclassify dog breeds based on the background of dog images. This happens when the backgrounds are correlated with other breeds in the training data, leading to misclassifications during test time. Previous SC benchmark datasets suffer from varying issues, e.g., over-saturation or only containing one-to-one (O2O) SCs, but no many-to-many (M2M) SCs arising between groups of spurious attributes and classes. In this paper, we present \benchmark-\{O2O, M2M\}-\{Easy, Medium, Hard\}, an image classification benchmark suite containing spurious correlations between classes and backgrounds. To create this dataset, we employ a text-to-image model to generate photo-realistic images and an image captioning model to filter out unsuitable ones. The resulting dataset is of high quality and contains approximately 152k images. Our experimental results demonstrate that state-of-the-art group robustness methods struggle with \benchmark, most notably on the Hard-splits with none of them getting over $70\%$ accuracy on the hardest split using a ResNet50 pretrained on ImageNet. By examining model misclassifications, we detect reliances on spurious backgrounds, demonstrating that our dataset provides a significant challenge.
\end{abstract}

\section{Introduction}
One of the reasons we have not deployed self-driving cars and autonomous kitchen robots everywhere is their catastrophic behavior in out-of-distribution (OOD) settings that differ from the training distribution \cite{d2020underspecification,geirhos2020shortcut}. To make models more robust to unseen test distributions, mitigating a classifier's reliance on spurious, non-causal features that are not essential to the true label has attracted lots of research interest \cite{groupdro, arjovsky2019invariant, survey, izmailov2022feature}. For example, classifiers trained on ImageNet \cite{deng2009imagenet} have been shown to rely on backgrounds \cite{xiao2020noise,singla2021salient,imagenet_spurious}, which are spuriously correlated with class labels but, by definition, not predictive of them. 

Recent work has focused substantially on developing new methods for addressing the spurious correlations (SCs) problem \cite{survey}, yet, studying and addressing the limitations of existing benchmarks remains underexplored. For example, the \emph{Waterbirds} \cite{groupdro}, and \emph{CelebA hair color} \cite{celeba} benchmarks remain among the most used benchmarks for the SC problem; yet, GroupDRO \cite{groupdro} achieves 90.5\% worst-group accuracy using group adjusted data with a ResNet50 pretrained on ImageNet.

\begin{figure}[t]
  \centering
  \begin{subfigure}[b]{0.48\linewidth}
    \centering
    \includegraphics[height=4.8cm]{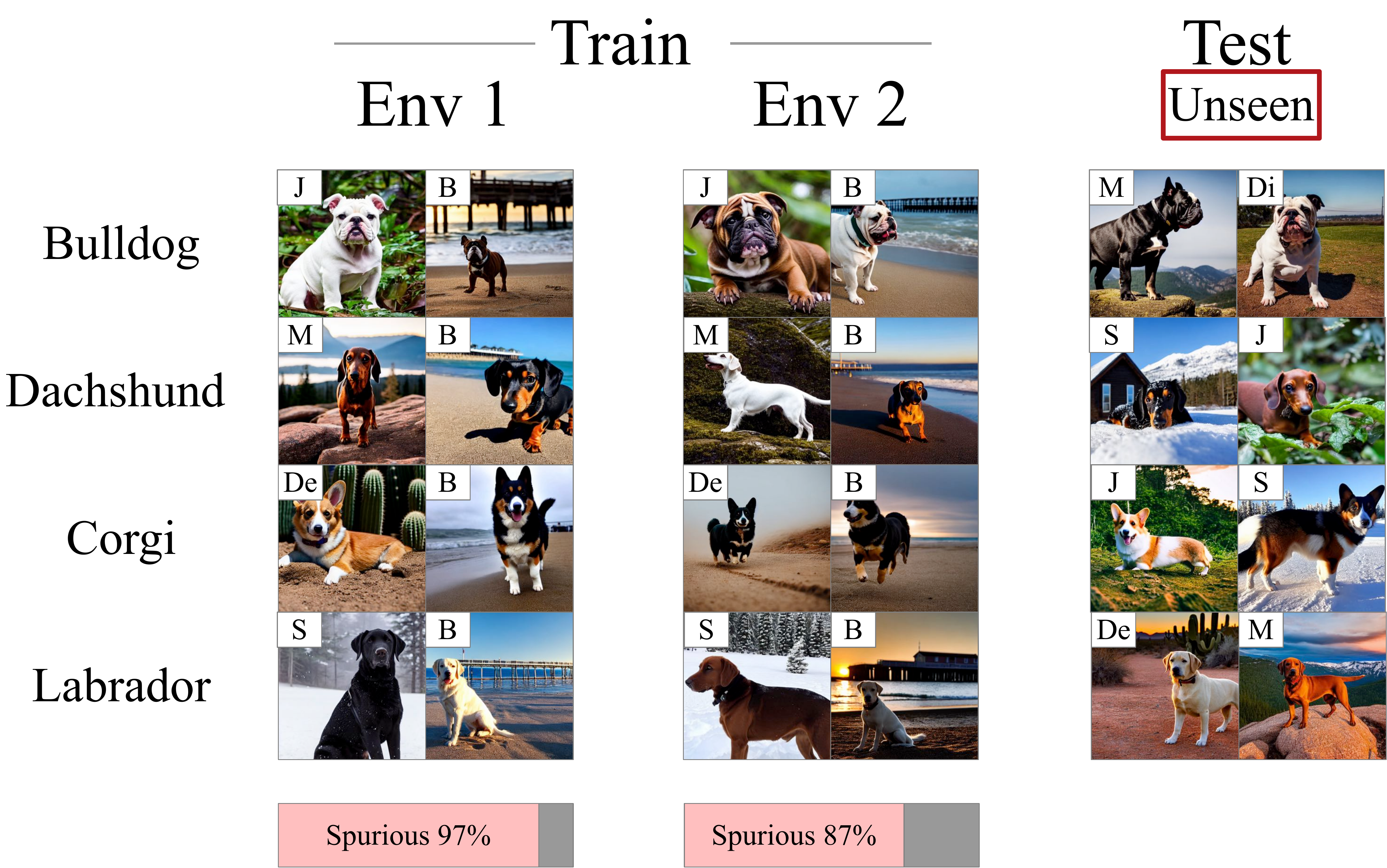}
    \caption{One-To-One (O2O)}
    \label{fig:o2o}
  \end{subfigure}
  \hfill
  \begin{subfigure}[b]{0.45\linewidth}
    \centering
    \includegraphics[height=4.8cm]{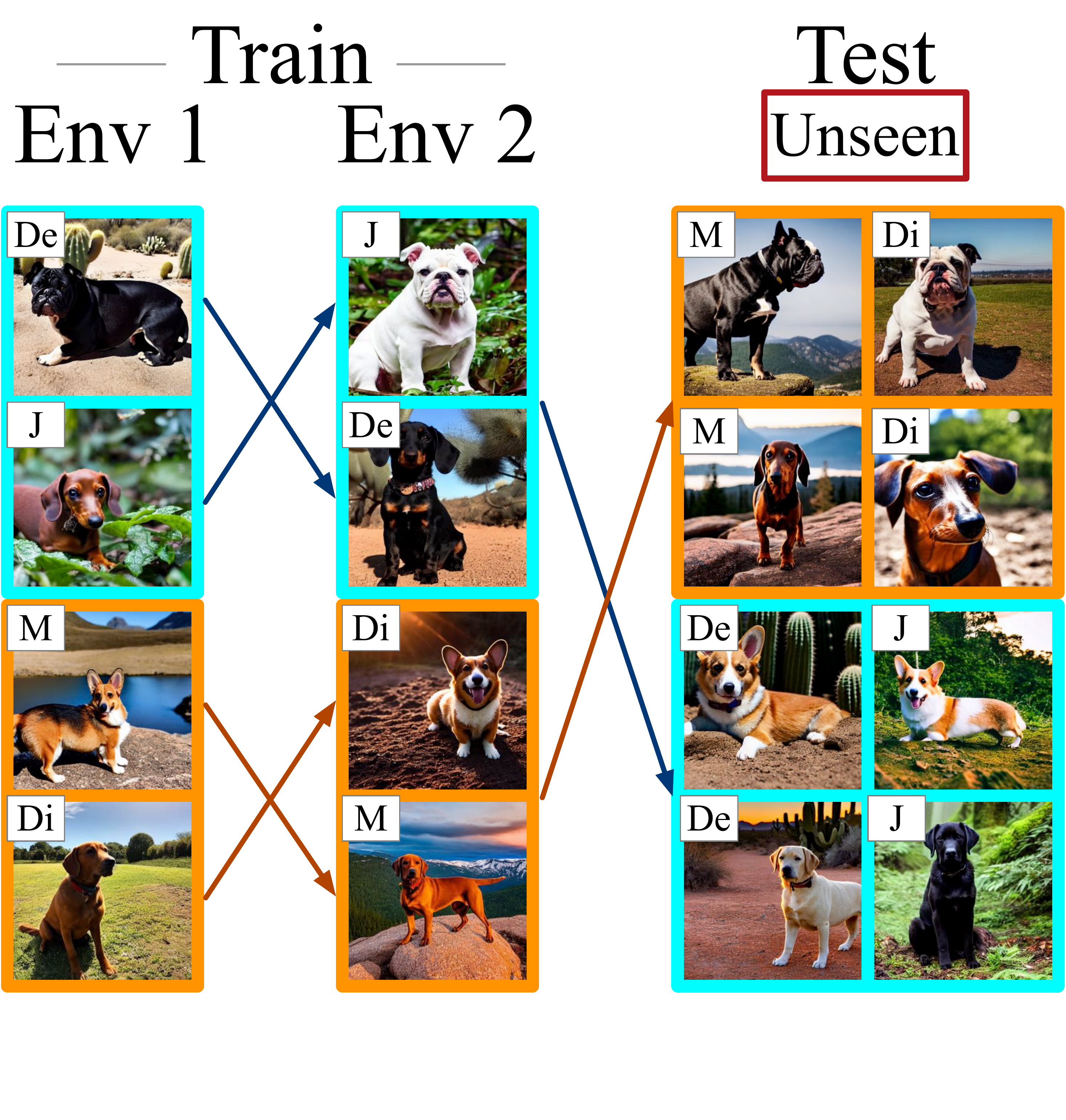}
    \caption{Many-To-Many (M2M)}
    \label{fig:explain_group}
  \end{subfigure}
  \caption{\textbf{\benchmark Challenges:} Letters on the images denote the background, and the bottom bar in \Cref{fig:o2o} indicates each class's proportion of the spurious background. In the O2O challenge, each class associates with a background during training, while the test data contains unseen combinations of class-background pairs. In the M2M challenge, a group of classes correlates with a group of backgrounds during training, but this correlation is reversed in the test data.}
  \label{fig:sc_challenge_overview}
\end{figure}

Another limitation of existing benchmarks is their sole focus on overly simplistic one-to-one (O2O) spurious correlations, where one spurious attribute correlates with one label. However, in reality, we often face \emph{many-to-many} (M2M) spurious correlations across groups of classes and backgrounds, which we formally introduce in this work. Imagine that during summer, we collect training data of two groups of two animal species (classes) from two groups of locations, e.g., a tundra and a forest in eastern Russia and a lake and mountain in western Russia. Each animal group correlates with a background group. In the upcoming winter, while looking for food, each group migrates, one going east and one going west, such that the animal groups have now exchanged locations. The spurious correlation has now been reversed in a way that cannot be matched from one class to one location.

\begin{figure}[t]
  \centering
  \begin{subfigure}[b]{0.45\linewidth}
    \includegraphics[height=2.4cm]{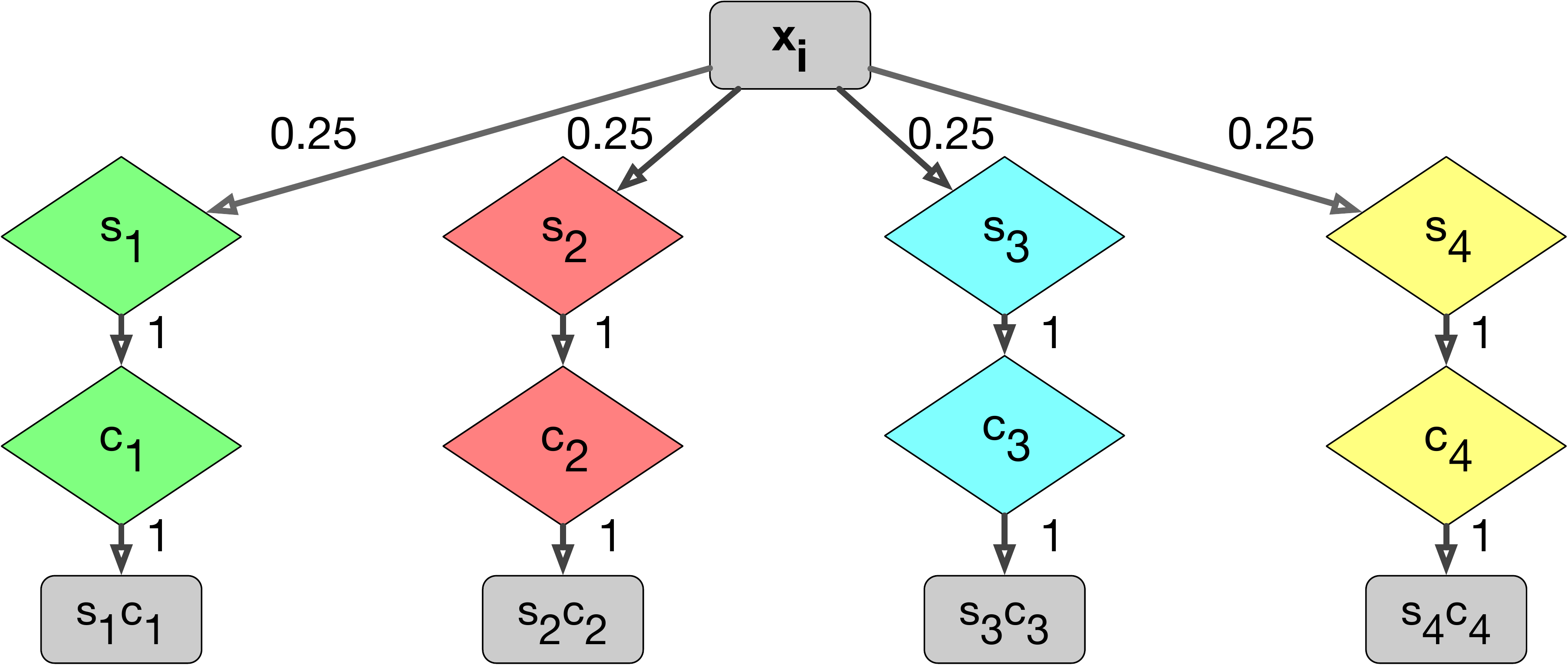}
    \caption{\textbf{O2O:} training data}
    \label{fig:o2o_train_dist}
  \end{subfigure}
  \hfill
  \begin{subfigure}[b]{0.45\linewidth}
    \includegraphics[height=2.4cm]{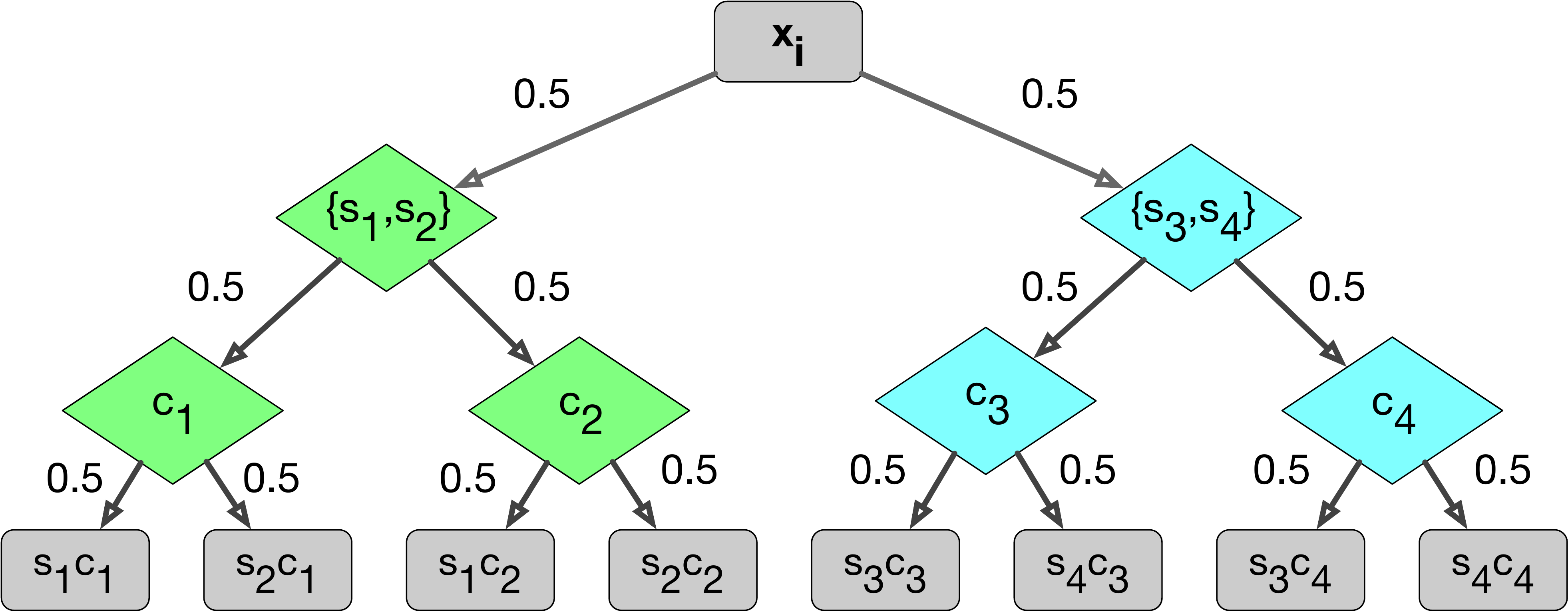}
    \caption{\textbf{M2M:} training data}
    \label{fig:m2m_train_dist}
  \end{subfigure}
  \\
  \begin{subfigure}[b]{0.45\linewidth}
    \includegraphics[height=2.4cm]{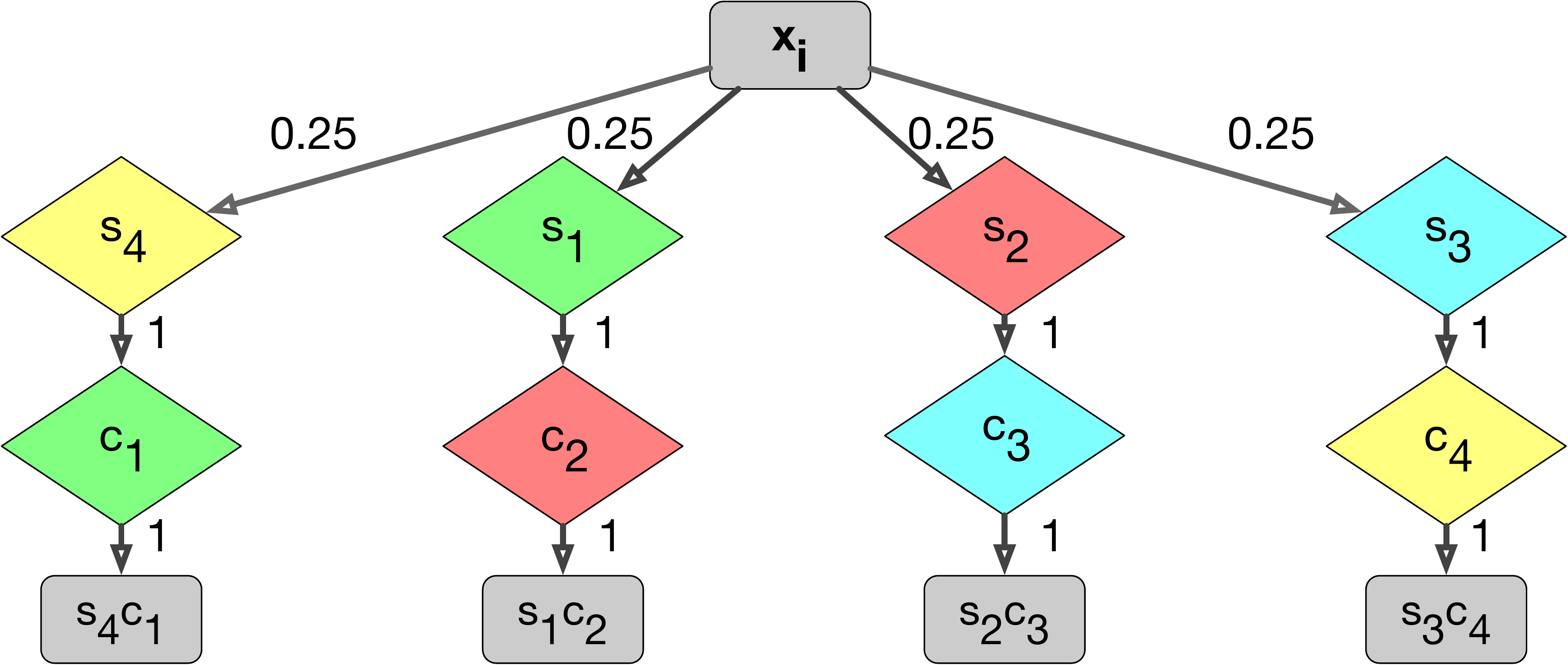}
    \caption{\textbf{O2O:} test data}
    \label{fig:o2o_test_dist}
  \end{subfigure}
  \hfill
  \begin{subfigure}[b]{0.45\linewidth}
    \includegraphics[height=2.4cm]{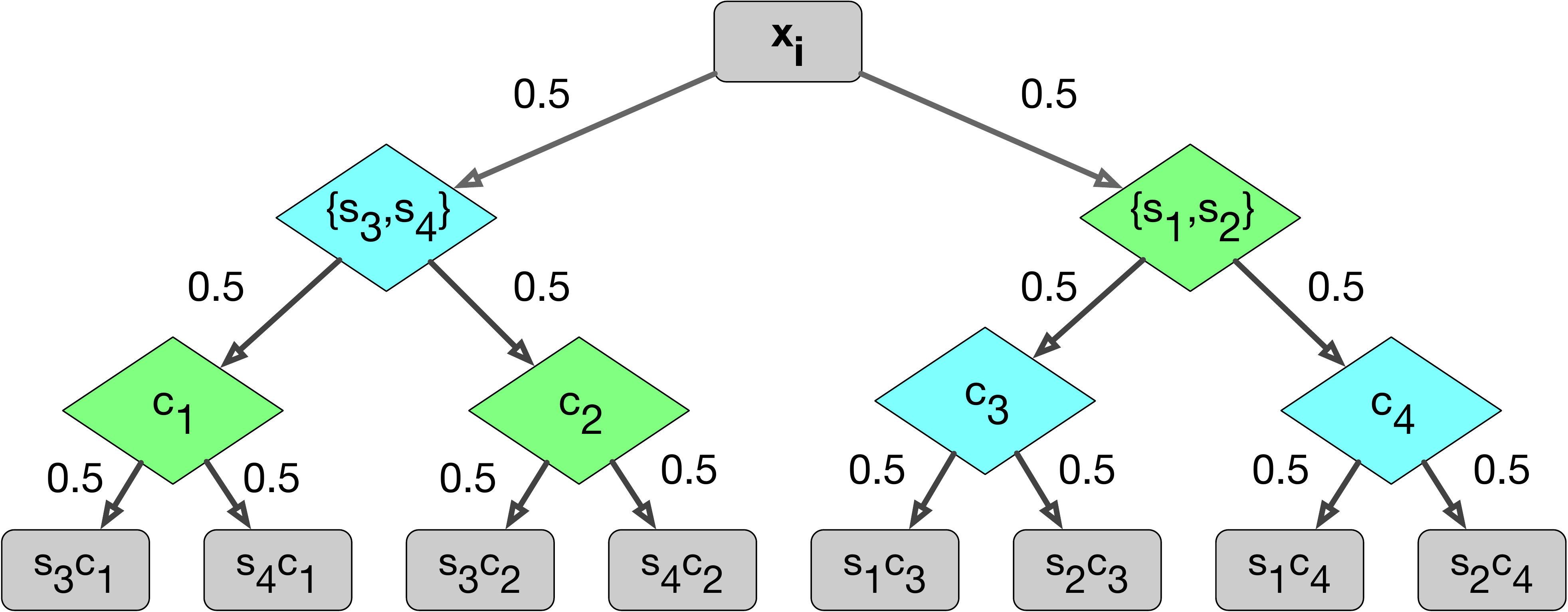}
    \caption{\textbf{M2M:} test data}
    \label{fig:m2m_test_dist}
  \end{subfigure}
  \caption{\textbf{Data distributions for our challenges:} $\vx_i$ is a random image sampled, each $s_i$ is a spurious attribute, and each $c_i$ is a class label. The edges indicate the probability that the sample $\vx_i$ has a given property, conditional on previous steps in the tree. The leaf nodes indicate the possible attribute-class combinations in the distribution. The colors emphasize the distribution shift in the test data.}
  \label{fig:four_images}
\end{figure}

While some benchmarks include multiple training environments with varying correlations \cite{wilds}, they do not test classification performance on reversed correlations during test time. Such M2M-SCs are \emph{not} an aggregation of O2O-SCs and cannot be expressed or decomposed in the form of the latter; they contain qualitatively different spurious structures, as shown in \Cref{fig:four_images}. To the best of our knowledge, this work is the first to conceptualize and instantiate M2M-SCs in image classification problems.

\paragraph{Contributions} We introduce \emph{\benchmark}-\{O2O, M2M\}-\{Easy, Medium, Hard\}, a suite of image classification datasets with O2O and M2M spurious correlations and three difficulty levels each. Recent work \cite{wiles2022discovering,lynch2022evaluating,vendrow2023dataset} has demonstrated a proof-of-concept to effectively discover spurious correlation failure cases in classifiers by leveraging off-the-shelf, large-scale, image-to-text models trained on vast amounts of data. Here, we take this view to the extreme and generate a novel benchmark with $152,064$ images of resolution $224 \times 224$, specifically targeted at the probing of classifiers' reliance on spurious correlations. 

Our experimental results demonstrate that state-of-the-art methods struggle with \benchmark, most notably on the \emph{Hard}-splits with $<70\%$ accuracy using ResNet50 pretrained on ImageNet. We probe a model's misclassifications and find further evidence for its reliance on spurious features. We also experiment with different model architectures, finding that while larger architectures can sometimes improve performance, the gains are inconsistent across methods, further raising the need for driving future research. 
\section{Existing Benchmarks}
\label{sec:existing_benchmarks}

We summarize the differences between \benchmark and related benchmarks in \Cref{tab:benchmark_table}. DomainBed \cite{domainbed} is a benchmark suite consisting of seven previously published datasets focused on domain generalization (DG), not on spurious correlations (excluding CMNIST, which we discuss separately). After careful hyper-parameter tuning, the authors find that ERM, not specifically designed for DG settings, as well as DG-specific methods, perform all about the same on average. They conjecture that these datasets may comprise an ill-posed challenge. For example, they raise the question of whether DG from a photo-realistic training environment to a cartoon test environment is even possible. In contrast, we follow the same rigorous hyper-parameter tuning procedure by \cite{domainbed} and observe stark differences among methods on \benchmark in \Cref{sec:results}, with ERM being the worst and $7.30\%$ points worse than the best method on average.
\begin{wraptable}{r}{0.4\textwidth}
\resizebox{.4\textwidth}{!}{
\begin{tabular}{|c|c|c|c|}
\hline
\textbf{Dataset} & \textbf{DG} & \textbf{O2O-SC} & \textbf{M2M-SC} \\
\toprule
CelebA-Hair Color \cite{celeba} & \xmark & \cmark & \xmark \\
\hline
Waterbirds \cite{groupdro} & \xmark & \cmark & \xmark \\
\hline
CMNIST \cite{arjovsky2019invariant} & \cmark & \cmark & \xmark \\
\hline
DomainBed${}^{*}$ \cite{domainbed} & \cmark & \xmark & \xmark \\ \hline
WILDS \cite{wilds} & \cmark & \xmark & \xmark \\
\hline
NICO \cite{zhang2023nico} & \cmark & \xmark & \xmark \\
\hline
MetaShift \cite{liang2022metashift} & \cmark & \xmark & \xmark \\
\hline
\textbf{\benchmark}  & \cmark & \cmark & \cmark \\
\bottomrule
\end{tabular}}
\caption{\textbf{Differences between \benchmark and other benchmarks}, according to whether they pose a Domain Generalization (DG), One-To-One- and/or Many-To-Many Spurious Correlations challenge.}
\label{tab:benchmark_table}
\vspace{-0.2cm}
\end{wraptable}

WILDS \cite{wilds}, NICO \cite{zhang2023nico} and MetaShift \cite{liang2022metashift} collect in-the-wild data and group data points with environment labels. However, these benchmarks do not induce \emph{explicit} spurious correlations between environments and labels. For example, WILDS-FMOW \cite{wilds,fmow} possesses a label shift between non-African and African regions; yet, the test images pose a domain generalization (DG) challenge (test images were taken several years later than training images) instead of reverting the spurious correlations observed in the training data. Waterbirds \cite{groupdro}, and CelebA hair color \cite{celeba,izmailov2022feature} are binary classification datsets including spurious correlations but without unseen  test domains (DG). Further, \citet{idrissi2022simple} illustrates that a simple class-balancing strategy alleviates most of their difficulty, while \benchmark is class-balanced from the beginning. ColorMNIST \cite{arjovsky2019invariant} includes spurious correlations and poses a DG problem. However, it is based on MNIST and, therefore, over-simplistic, i.e., it does not reflect real-world spurious correlations involving complex background features, such as the ones found in ImageNet \cite{singla2021salient,imagenet_spurious}.

None of the above benchmarks include explicit training and test environments for M2M-SCs. 

\section{Benchmark Desiderata}
\label{sec:desiderata}
Motivated by the shortcomings of previous benchmarks discussed in \Cref{sec:existing_benchmarks}, we want first to posit some general desiderata that an improved benchmark dataset would satisfy. Next, we motivate and formalize the two types of spurious correlations we aim to study. 
\subsection{Six Desiderata}
\orange{1. Photo-realism} unlike datasets containing cartoon/sketch images \cite{domainbed} or image corruptions \cite{corruptions}, which are known to conflict with current backbone network architectures \cite{geirhos2018imagenet,geirhos2018generalisation,hermann2020origins}, possibly confounding the evaluation of OOD algorithms. \red{2. Non-binary classification problem}, to minimize accidentally correct classifications achieved by chance. \blue{3. Inter-class homogeneity and intra-class heterogeneity}, i.e., low variability \emph{between} and high variability \emph{within} classes, to minimize the margins of the decision boundaries inside the data manifold \cite{pml1Book}. This desideratum ensures that the classification problem is non-trivial. \green{4. High-fidelity backgrounds} including complex features to reflect realistic conditions typically faced in the wild instead of monotone or entirely removed backgrounds \cite{xiao2020noise}. 
\purple{5. Access to multiple training environments}, i.e., the conditions of the \emph{Domain Generalization} problem \cite{domainbed}, which allow us to learn domain invariances, such that classifiers can perform well in novel test domains. \cyan{6. Multiple difficulty levels}, so future work can study cost trade-offs. For example, we may be willing to budget higher computational costs for methods succeeding on difficult datasets than those that succeed only on easy ones.

\subsection{Spurious Correlations (One-To-One)}

Here, we provide some intuition and discuss the conditions for a (one-to-one) spurious correlation (SC). We define a correlated, non-causal feature as a feature that frequently occurs with a class but does not cause the appearance of the class (nor vice versa). We abuse the term ``correlated'' as it is commonly used by previous work, but we consider non-linear relationships between two random variables too. Further, we call correlated features \emph{spurious} if the classifier perceives them as a feature of the correlated class.

Next, we want to define a \emph{challenge} that allows us to evaluate a classifier's harmful reliance on spurious features. Spurious features are not always harmful; even humans use context information to make decisions \cite{geirhos2020shortcut}. However, a spurious feature becomes harmful if it alone is sufficient to trigger the prediction of a particular class without the class object being present in the image \cite{imagenet_spurious}. 

To evaluate a classifier w.r.t. such harmful predictions, we evaluate its performance when the spurious correlations are reverted. The simplest setting is when a positive/negative correlation exists between one background variable and one label in the training/test environment. We formalize this setup as follows. 

\begin{challenge}{O2O-SC Challenge} 
Let $p(\rmX, S, C)$ be a distribution over images $\rmX \in \R^D$, spurious attributes $S \in \mathcal{S} = \{s_1, \dots, s_K\}$, and labels $C \in \mathcal{C} = \{c_1, \dots, c_P \}$.
Given $\ptrain \neq \ptest$, and $K = P$ it holds that for $i \in [K]$, 
\begin{align}
\corr_{\ptrain}\left(\ind(S = s_i), \ind(C=c_i)\right) > 0, \; \corr_{\ptest}\left(\ind(S = s_i), \ind(C = c_i)\right) < 0.
\end{align}
where the indicator function $\ind(X=x )$ is non-zero when the \emph{variable} $X$ equals the \emph{value} $x$.
\end{challenge}

By one-to-one (O2O) SC, we refer to a setting in which pair-wise SCs between spurious features $S$ (also called \emph{style} variable \cite{survey}) and labels $C$ exist within training environments, which then differ in the test environment, as illustrated in \Cref{fig:o2o}.

\subsection{Many-To-Many Spurious Correlations}
In this subsection, we conceptualize Many-To-Many (M2M) SCs, where the SCs hold over disjoint groups of spurious attributes and classes. For instance, in \cref{fig:explain_group}, each class from the class group $\{$\emph{Bulldog}, \emph{Dachshund}$\}$ is observed with each background from the group $\{$\emph{Desert}, \emph{Jungle}$\}$ in equal proportion in the training data.

\begin{challenge}{M2M-SC Challenge}
Consider $p(\rmX, S, C)$ defined in the O2O-SC Challenge. We further assume the existence of partitions $\mathcal{S} = \mathcal{S}_1 \dot\cup \mathcal{S}_2$ and $\mathcal{C} = \mathcal{C}_1 \dot\cup \mathcal{C}_2$.
Given $\ptrain, \ptest$, it holds that for $j \in \{1,2\}$
\begin{align}
\corr_{\ptrain}\left(\ind(S \in \mathcal{S}_{j}), \ind(C \in \mathcal{C}_{j})\right) = 1 , \corr_{\ptest} \left(\ind(S \in \mathcal{S}_{j}), \ind(C \in \mathcal{C}_{j})\right) = -1.
\end{align}
\end{challenge}

\Cref{fig:four_images} shows an example of how to construct M2M-SCs, which contain richer spurious structures, following an \emph{hierarchy} of the class groups correlating with spurious attribute groups. As we will see later in \Cref{pg:m2msc_construction}, the data-generating processes we instantiate for each challenge differ qualitatively.

\section{The \benchmark Challenge}
In this section, we instantiate the desiderata introduced in \Cref{sec:desiderata} by presenting \emph{\benchmark}, a synthetic image classification dataset containing images of four dog breeds (classes) in six background locations (spurious attributes). 

\begin{figure*}
    \centering
    \includegraphics[width=0.8\linewidth]{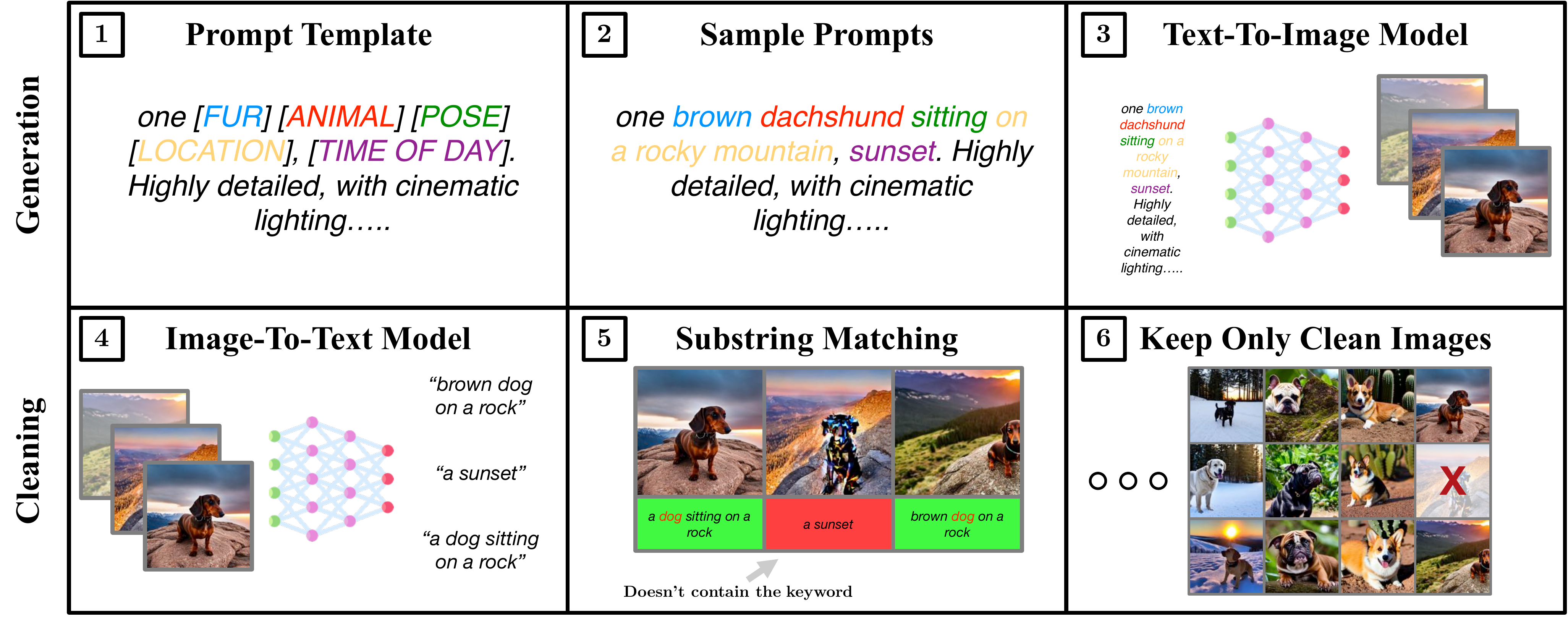}
    \caption{\textbf{\benchmark Pipeline:} We leverage text-to-image models for data generation and image-to-text models for cleaning.}
    \label{fig:generation_pipeline}
\end{figure*}

\subsection{Dataset Construction}
\Cref{fig:generation_pipeline} summarizes the dataset construction pipeline, which we now discuss in more detail. The main idea is to leverage recently proposed text-to-image models \cite{rombach2022high} for photo-realistic image generation and image-to-text models \cite{nlp_connect_2022} for filtering out low-quality images.

A \textbf{prompt template} allows us to define high-level factors of variation, akin to a causal model \cite{survey} of the data-generating process. 
We then \textbf{sample prompts} by filling in randomly sampled values for these high-level factors. 
The \textbf{text-to-image model} generates images given a sampled prompt; we use \emph{Stable Diffusion v1.4} \cite{rombach2022high}.
We pass the raw, generated images to an \textbf{image-to-text (I2T) model} to extract a concise description; here, we use the ViT-GPT2 image captioning model \cite{nlp_connect_2022}.
We perform a form of \textbf{substring matching} by checking whether important keywords are present in the caption, e.g., \emph{``dog''}. This step avoids including images without class objects, which we sometimes observed due to the T2T model ignoring parts of the input prompt.  
We \textbf{keep only ``clean'' images} whose captions include important keywords.

\subsection{Satisfying Desiderata}
To ensure \orange{photorealism}, we generate images using \emph{Stable Diffusion v1.4} \cite{rombach2022high}, trained on a large-scale real-world image dataset \cite{laion5b}, while carefully filtering out images without detectable class objects. We construct a 4-way classification problem to reduce the probability of accidentally-correct classifications compared to a \red{binary classification problem} (e.g., CelebA hair color prediction or Waterbirds). Next, we chose dog breeds to reduce \blue{inter-class variance}, inspired by the difference in classification difficulty between Imagenette (easily classified objects) \cite{Howard_Imagenette_2019}, and ImageWoof \cite{Howard_Imagewoof_2019} (dog breeds), two datasets based on subsets of ImageNet \cite{deng2009imagenet}. We increase \blue{intra-class variance} by adding animal poses to the prompt template. 

We add ``\emph{[location] [time of day]}'' variables to the prompt template to ensure \green{diverse backgrounds}, and select six combinations after careful experimentation with dozens of possible combinations, abandoning over-simplistic ones. Our final prompt template takes the form 
``\emph{one [fur] [animal] [pose] [location], [time of day]. highly detailed, with cinematic lighting, 4k resolution, beautiful composition, hyperrealistic, trending, cinematic, masterpiece, close up}'',
and there are 72 possible combinations. 
The variables [location]/[animal] correspond to spurious backgrounds/labels for a specific background-class combination.
The other variables take the following values: \emph{fur:} black, brown, white, [empty]; \emph{pose:} sitting, running, [empty]; \emph{time of day:} pale sunrise, sunset, rainy day, foggy day, bright sunny day, bright sunny day

To construct \purple{multiple training environments}, we randomly sample from a set of background-class combinations, which we further group by their \cyan{difficulty level into \emph{easy}, \emph{medium}, and \emph{hard}}. We construct two datasets for each SC type with $3168$ images per background-class combination, thus $2 \; \text{SC types} \; \times 4 \; \text{environments} \; \times 6 \; \text{difficulties} \; \times 3168  = 152,064$ images in total.

\paragraph{O2O-SC Challenge} \label{para:o2o_implementation}
We select combinations such that each class is observed with two backgrounds, spurious $b^{\text{sp}}$ and generic $b^{\text{ge}}$. For all images with class label $c_i$ in the training data, $\mu$\% of them have the spurious background $b^{\text{sp}}_i$ and $(100-\mu)$\% of them have the generic background $b^{\text{ge}}$. Importantly, each spurious background is observed with only one class ($\ptrain(b^{\text{sp}}_i \mid c_j) = 1$ if $i=j$ and $0$ for $i \neq j$), while the generic background is observed for all classes with equal proportion. We train on two separate environments (with distinct data) that differ in their $\mu$ values. Thus, the change in this proportion should serve as a signal to a robustness-motivated optimization algorithm (e.g. IRM \cite{arjovsky2019invariant}, GroupDRO \cite{groupdro} etc.) that the correlation is spurious. 

For instance, in \cref{fig:o2o}, training environment 1, 97\% of the \emph{Bulldog} images have spurious \emph{Jungle} backgrounds, while 3\% have generic \emph{Beach} backgrounds. The spurious background changes depending on the class, but the relative proportions between each trio $c_i, b_i^{\text{sp}}$ and $b_i^{\text{ge}}$ are the same. Notably, in training environment 2, the proportions change to 87\% and 13\% split of spurious and generic backgrounds.

\begin{table*}
        \centering
        \resizebox{\textwidth}{!}{
        \begin{tabular}{|c|ccc|ccc|ccc|}
        \toprule
        \bf Class & \bf Train Env 1 & \bf Train Env 2 & \bf Test & \bf Train Env 1 & \bf Train Env 2 & \bf Test & \bf Train Env 1 & \bf Train Env 2 & \bf Test   \\
        \midrule
        & \multicolumn{3}{c|}{\bf O2O-Easy} &  \multicolumn{3}{c|}{\bf O2O-Medium}  & \multicolumn{3}{c|}{\bf O2O-Hard}\\
        \midrule
        Bulldog & 97\% De 3\% B & 87\% De 13\% B & 100\% Di 
        & 97\% M 3\% De & 87\% M 13\% De & 100\% J 
        & 97\% J 3\% B & 87\% J 13\% B & 100\% M \\
        
        Dachshund & 97\% J 3\% B & 87\% J 13\% B & 100\% S 
        & 97\% B 3\% De & 87\% B 13\% De & 100\% Di 
        & 97\% M 3\% B & 87\% M 13\% B & 100\% S \\

        Labrador & 97\% Di 3\% B & 87\% Di 13\% B & 100\% De 
        & 97\% Di 3\% De & 87\% Di 13\% De & 100\% B 
        & 97\% S 3\% B & 87\% S 13\% B & 100\% De \\

        Corgi & 97\% S 3\% B & 87\% S 13\% B & 100\% J 
        & 97\% J 3\% De & 87\% J 13\% De & 100\% S 
        & 97\% De 3\% B & 87\% De 13\% B & 100\% J \\

        \midrule
        & \multicolumn{3}{c|}{\bf M2M-Easy} &  \multicolumn{3}{c|}{\bf M2M-Medium}  & \multicolumn{3}{c|}{\bf M2M-Hard}\\
        \midrule
        
        Bulldog & 100\% Di & 100\% J & 50\% S 50\% B 
        & 100\% De & 100\% M & 50\% Di 50\% J 
        & 100\% B & 100\% S & 50\% De 50\% M  \\
        
        Dachshund & 100\% J & 100\% Di & 50\% S 50\% B 
        & 100\% M & 100\% De & 50\% Di 50\% J 
        & 100\% B & 100\% S & 50\% De 50\% M \\

        Labrador & 100\% S & 100\% B & 50\% Di 50\% J 
        & 100\% Di & 100\% J & 50\% De 50\% M 
        & 100\% M & 100\% De & 50\% B 50\% S \\

        Corgi & 100\% B & 100\% S & 50\% Di 50\% J 
        & 100\% J & 100\% Di & 50\% De 50\% M 
        & 100\% M & 100\% De & 50\% B 50\% S  \\

        \bottomrule
        \end{tabular}}

     \caption{\textbf{Proportions of Spurious Backgrounds By Class and Environment.} Backgrounds include: Beach (B), Desert (De), Dirt (Di), Jungle (J), Mountain (M), Snow (S).}
     \label{tab:data_combinations}
\end{table*}

\paragraph{M2M-SC Challenge} 
\label{pg:m2msc_construction}

First, we construct disjoint background and class groups $\gS_1, \gS_2, \gC_1, \gC_2$, each with two elements. Then, we select background-class combinations for the training data such that for each class $c \in \gC_i$, we pick a combination $(s,b)$ for each $s \in \gS_i$. Second, we introduce two environments as shown in \cref{fig:explain_group}.

The difference between \emph{Easy}, \emph{Medium}, and \emph{Hard} is in the splits between the available combinations of backgrounds and classes. We determined the final splits based on the empirical performance of a standard ERM classifier \Cref{sec:methods}). There are no theoretical reasons we present to explain the difference in difficulty in the challenges.

\paragraph{Data combination selection} Each spurious correlation (O2O, M2M) comes with three data combinations (Easy, Medium,Hard) and a test environment. We selected one data combination while experimenting with ways to build spurious correlations. Those are \benchmark-\{O2O\}-\{Easy\} and \benchmark-\{M2M\}-\{Hard\} which we used for early experiments and fine-tuning. To find the remaining data combinations for \benchmark-\{O2O\}-\{Medium, Hard\} and \benchmark-\{M2M\}-\{Easy, Medium\}, we performed a random search within a space of possible data combinations for 20 random trials on O2O and M2M each. Full details on the final data combinations are here: \cref{tab:data_combinations}.

\section{Experiments}
\label{sec:experiments}

\begin{table*}[t]
        \centering
        \resizebox{\textwidth}{!}{
        \begin{tabular}{c|ccc|ccc|c}
        \toprule
        \bf \multirow{2}{*}{Method} & \multicolumn{3}{c|}{\bf One-To-One SC} &  \multicolumn{3}{c|}{\bf Many-To-Many SC}  & \bf \multirow{2}{*}{Average}\\
        \cmidrule{2-7}
        & \bf Easy & \bf Medium & \bf Hard & \bf Easy & \bf Medium & \bf Hard &   \\
        \midrule
        \multicolumn{8}{c}{\textbf{ResNet18} pre-trained on ImageNet} \\
        \midrule
        ERM \cite{vapnik1991principles}& $72.15\%_{\pm 0.03}$ & $69.85\%_{\pm 0.01}$ & $65.65\%_{\pm 0.04}$ & $72.51\%_{\pm 0.05}$ & $51.36\%_{\pm 0.04}$ & $47.02\%_{\pm 0.01}$ & $63.09\%$\\
        GroupDRO \cite{groupdro} & $68.72\%_{\pm 0.02}$ & $71.87\%_{\pm 0.01}$ & $60.90\%_{\pm 0.03}$ & $74.82\%_{\pm 0.04}$ & $52.06\%_{\pm 0.03}$ & $52.79\%_{\pm 0.03}$ & $63.53\%$\\
        IRM \cite{arjovsky2019invariant} & $71.26\%_{\pm 0.02}$ & $68.18\%_{\pm 0.02}$ & $63.78\%_{\pm 0.03}$ & $73.28\%_{\pm 0.04}$ & $42.43\%_{\pm 0.07}$ & $44.51\%_{\pm 0.06}$ & $60.57\%$ \\
        CORAL \cite{sun2016deep} & $83.85\%_{\pm 0.01}$ & $73.96\%_{\pm 0.01}$ & $72.18\%_{\pm 0.03}$ & $79.91\%_{\pm 0.00}$ & $58.09\%_{\pm 0.01}$ & $56.51\%_{\pm 0.03}$ & $70.75\%$ \\
        CausIRL \cite{chevalley2022invariant} & $\mathbf{84.21\%_{\pm 0.01}}$ & $73.45\%_{\pm 0.02}$ & $71.20\%_{\pm 0.02}$ & $81.21\%_{\pm 0.01}$ & $56.79\%_{\pm 0.01}$ & $56.31\%_{\pm 0.03}$ & $70.52\%$\\
        MMD-AAE \cite{li2018domain} & $82.92\%_{\pm 0.03}$ & $\mathbf{74.09\%_{\pm 0.03}}$ & $\mathbf{72.60\%_{\pm 0.05}}$ & $\mathbf{83.45\%_{\pm 0.01}}$ & $\mathbf{60.27\%_{\pm 0.03}}$ & $\mathbf{58.26\%_{\pm 0.00}}$ & $\mathbf{71.93\%}$ \\
        \midrule
        \multicolumn{8}{c}{\textbf{ResNet50} pre-trained on ImageNet} \\
        \midrule
        ERM \cite{vapnik1991principles}& $77.49\%_{\pm 0.05}$ & $76.60\%_{\pm 0.02}$ & $71.32\%_{\pm 0.09}$ & $83.80\%_{\pm 0.01}$ & $53.05\%_{\pm 0.03}$ & $58.70\%_{\pm 0.04}$ & $70.16\%$\\
        GroupDRO \cite{groupdro} & $80.58\%_{\pm 0.01}$ & $75.96\%_{\pm 0.02}$ & $76.99\%_{\pm 0.03}$ & $79.96\%_{\pm 0.03}$ & $61.01\%_{\pm 0.05}$ & $60.86\%_{\pm 0.02}$ & $72.56\%$\\
        IRM \cite{arjovsky2019invariant} & $75.45\%_{\pm 0.03}$ & $76.39\%_{\pm 0.02}$ & $74.90\%_{\pm 0.01}$ & $76.15\%_{\pm 0.03}$ & $\mathbf{67.82\%_{\pm 0.04}}$ & $60.93\%_{\pm 0.01}$ & $71.94\%$\\
        CORAL \cite{sun2016deep} & $\mathbf{89.66\%_{\pm 0.01}}$ & $\mathbf{81.05\%_{\pm 0.01}}$ & $79.65\%_{\pm 0.02}$ & $81.26\%_{\pm 0.02}$ & $65.18\%_{\pm 0.05}$ & $67.97\%_{\pm 0.01}$ & $\mathbf{77.46\%}$\\
        CausIRL \cite{chevalley2022invariant} & $89.32\%_{\pm 0.01}$ & $78.64\%_{\pm 0.01}$ & $\mathbf{80.40\%_{\pm 0.01}}$ & $\mathbf{85.76\%_{\pm 0.01}}$ & $63.15\%_{\pm 0.03}$ & $\mathbf{68.93\%_{\pm 0.01}}$ & $77.20\%$\\
        MMD-AAE \cite{li2018domain} & $78.81\%_{\pm 0.02}$ & $75.33\%_{\pm 0.03}$ & $72.66\%_{\pm 0.01}$ & $80.55\%_{\pm 0.02}$ & $59.43\%_{\pm 0.04}$ & $54.39\%_{\pm 0.05}$ & $70.20\%$\\
        \bottomrule
        \end{tabular}}
     \caption{\textbf{Results for \benchmark-\{O2O,M2M\}-\{Easy, Medium, Hard\}} } 
     \label{tab:results}
\end{table*}

        


In this section, we evaluate various group robustness methods on \benchmark and discuss the results.
\paragraph{Methods} \label{sec:methods}
The field of worst-group-accuracy optimization is thriving with a plethora of proposed methods, making it impractical to compare all available methods. We choose the following six popular methods and their \domainbed implementation \cite{domainbed}. \textbf{ERM} \cite{erm} refers to the canonical, average-accuracy-optimization procedure, where we treat all groups identically and ignore group labels, not targeting to improve the worst group performance.
 \textbf{CORAL} \cite{sun2016deep} penalizes differences in the first and second moment of the feature distributions of each group. 
 \textbf{IRM} \cite{arjovsky2019invariant} is a causality-inspired \cite{survey} invariance-learning method, which penalizes feature distributions that have different optimal linear classifiers for each group.
 \textbf{CausIRL} \cite{chevalley2022invariant} is another causally-motivated algorithm for learning invariances, whose penalty considers only one distance between mixtures of latent features coming from different domains.
 \textbf{GroupDRO} \cite{groupdro} uses Group-Distributional Robust Optimization to explicitly minimize the worst group loss instead of the average loss.
 \textbf{MMD-AAE} \cite{li2018domain} penalizes distances between feature distributions of groups via the maximum mean discrepancy (MMD) and learning an adversarial auto-encoder (AAE).

\paragraph{Hyper-parameter tuning} We follow the hyper-parameter tuning process used in \domainbed \cite{domainbed} with a minor modification. We keep the dropout rate ($0.1$) and the batch size fixed (128 for ResNets and 64 for ViTs) because we found them to have only a very marginal impact on the performance. We tune the learning rate and weight decay on ERM with a random search of 20 random trials. For all other methods, we further tune their method-specific hyper-parameters with a search of 10 random trials. We perform model selection based on the training domain validation accuracy of a subset of the training data. 
We reuse the hyper-parameters found for \benchmark-\{O2O\}-\{Easy\} and \benchmark-\{M2M\}-\{Hard\} on \benchmark-\{O2O\}-\{Medium, Hard\} and \benchmark-\{M2M\}-\{Easy, Medium\}, respectively. We also initially explored the ViT \cite{vit} architecture, with results shown in \Cref{sec:architectures}. Due to its poor performance, we chose to focus on ResNet50 results. 

\paragraph{Evaluation} We evaluate the classifiers on a test environment where the SCs present during training change, as described in \Cref{tab:data_combinations}. For O2O, multiple ways exist to choose a test data combination; we evaluate one of them as selected using a random search process. In M2M, because there are only two class groups and two background groups, we only need to swap them as seen in \cref{fig:explain_group}.

\subsection{Results}
\label{sec:results}

For our main results in \Cref{tab:benchmark_table}, we fine-tune a ResNet18 \cite{resnet} model pre-trained on ImageNet, following previous work on domain generalization \cite{dou2019domain,li2019episodic,domainbed}. We also fine-tune larger models later in \Cref{sec:architectures}. 

There is a clear hierarchy among methods; the top three listed methods perform significantly worse than the bottom three when using a ResNet18. The same hierarchy is preserved when using a ResNet50 except for the MMD-AAE method which gets worse after increasing the model size. This stimulating new result contributes to the debate whether, for a fixed architecture, most robustness methods perform about the same \cite{domainbed} or not \cite{wiles2021fine}. 
The performances of most methods get consistently worse as the challenge becomes harder.
Most often, the data splits of our newly formalized M2M-SC are significantly more challenging than the O2O splits, most notably \emph{M2M-\{Hard, Medium\}}. We conjecture that there is a strong need for new methods targeting such. 
\{ERM, GroupDRO\} and \{CORAL, CausIRL\} perform about the same, despite much different robustness regularization. 
All methods consistently achieve 98-99\% in-distribution test performance (not shown in \Cref{tab:benchmark_table} to save space) despite differences in OOD performance.
IRM performs worst on average, with a significant difference of 11.36\% compared to the best method (MMD-AAE). This result empirically confirms recent warnings about IRM's effectiveness \cite{rosenfeld2020risks,kamath2021does}.

\subsection{Misclassification analysis}
In \Cref{sec:results}, we learned that ERM performs particularly poorly on both hard challenges. Now, we want to investigate why by examining some of the misclassifications.
\begin{figure*}[t]
    \centering
    \includegraphics[width=\textwidth]{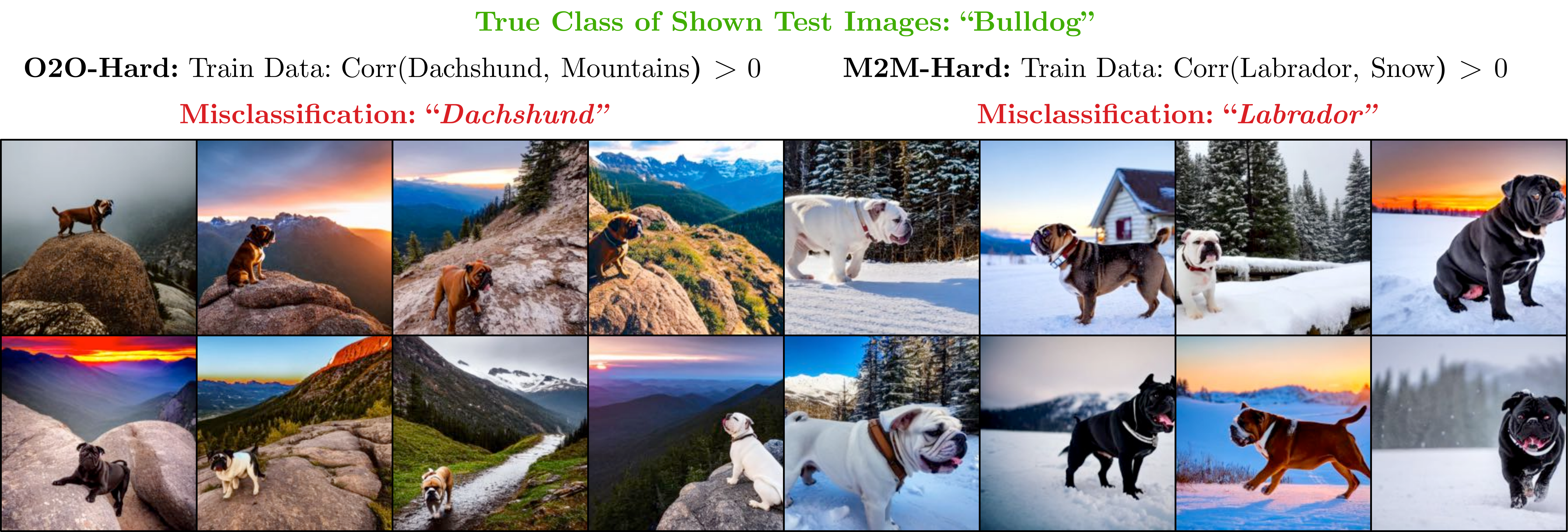}
    \caption{\textbf{ERM misclassifications due to spurious correlations.} The shown test images correspond to the class \emph{``Bulldog''} with spurious backgrounds \emph{``Mountains''} in the O2O-Hard (left) and \emph{``Snow''} in the M2M-Hard (right) challenge.  }
    \label{fig:misclassifications}
\end{figure*}
For example, we observe in \Cref{fig:misclassifications} that on the test set, the class \emph{``Bulldog''} is misclassified as the classes whose most common training set background is the same as \emph{``Bulldog''}'s test backgrounds. 

Note that for all classes and in all data groups, both training and test environments, the number of data points per class is always balanced; rendering methods like \emph{Subsampling large classes} \cite{idrissi2022simple}, which achieve state-of-the-art performance on other SC benchmarks, inapplicable. Hence, we conjecture that despite balanced classes, the model heavily relies on the spurious features of the \emph{``Mountains''} and \emph{``Snow''} backgrounds.
\begin{figure}[h]
  \begin{center}
    \begin{subfigure}{0.25\textwidth}
        \centering
        \includegraphics[width=\textwidth]{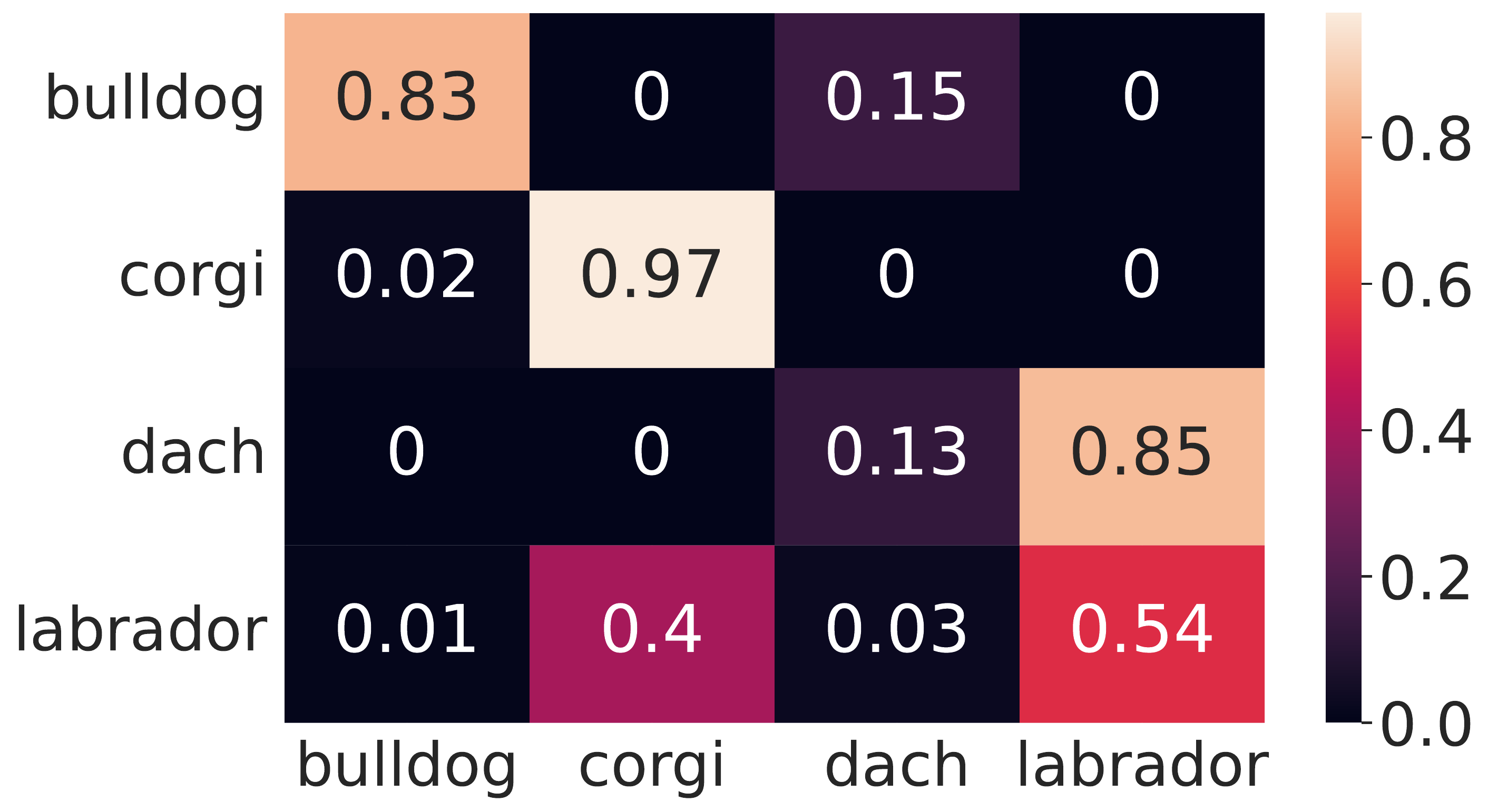}
        \caption{\textbf{O2O-Hard}} \label{confusion_o2o}
    \end{subfigure}    \begin{subfigure}{0.25\textwidth}
        \centering
        \includegraphics[width=\textwidth]{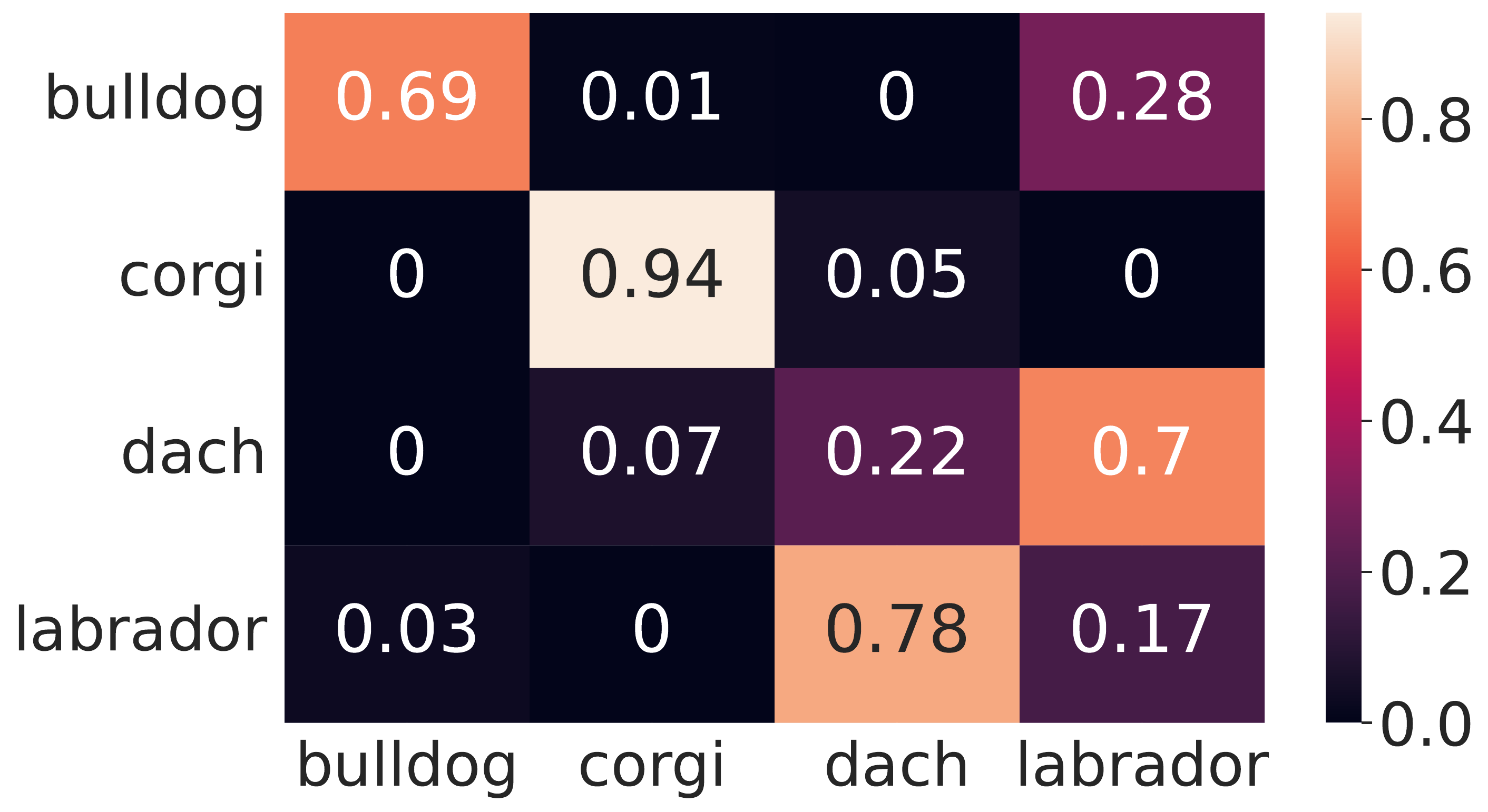}
        \caption{\textbf{{M2M-Hard}}}
    \end{subfigure}
  \end{center}
  \caption{\textbf{Confusion matrices for ERM models.} $X$-axis: predictions; $Y$-axis: true labels.}
  \label{fig:confusionmatrix}
\end{figure}

We further corroborate that claim by examining the model's confusion matrix in \Cref{fig:confusionmatrix}. For example, \Cref{confusion_o2o} shows the highest non-diagonal value for actual \emph{``Dachshund''} images being wrongly classified as \emph{``Labrador''}. We conjecture the reason being that in O2O-Hard, the background of \emph{``Dachshund''} in the test set is \emph{``Snow''}, which is the most common background of the training images of \emph{``Labrador''}, as shown in \Cref{tab:data_combinations}. 

\section{Related Work}

\paragraph{Failure Mode Analyses of Distribution Shifts}
\citet{nagarajan2020understanding} analyze distribution shift failure modes and find that two different kinds of shifts can cause issues: \emph{geometric} and \emph{statistical} skew. Geometric skew occurs when there is an imbalance between groups of types of data points (i.e., data points from different environments) which induces a spurious correlation and leads to misclassification when the balance of groups changes. This understanding has motivated simply 
removing data points from the training data to balance between groups of data points \cite{arjovsky2022throwing}. In contrast, we study two particular types of SCs, which persist in degenerating generalization performance despite perfect balances of classes among groups.

\paragraph{Causality} The theory of causation provides another perspective on the sources and possible mitigations of spurious correlations \cite{peters2017elements,survey}. Namely, we can formalize environment-specific data as samples from different interventional distributions, which keep the influence of variables not affected by the corresponding interventions invariant. This perspective has motivated several invariance-learning methods that make causal assumptions on the data-generating process \cite{arjovsky2019invariant,mahajan2021domain,mao2022causal}. The field of treatment effect estimation also deals with mitigating spurious correlations from observational data \cite{chernozhukov2018double,kunzel2019metalearners,kaddour2021causal,nie2021quasi}.

\section{Limitations and Future Work} Instantiating our desiderata with \textbf{non-background} spurious attributes. For example, \citet{imagenet_spurious} find that in the ImageNet \cite{deng2009imagenet} dataset, the class \emph{``Hard Disc''} is spuriously correlated with \emph{``label''}, however, \emph{``label''} is not a background feature but rather part of the classification object. Instantiating our desiderata for \textbf{other data modalities}, e.g., text classification, leveraging the text generation capabilities of large language models \cite{brown2020language}.
Evaluating \textbf{more generalization techniques} on \benchmark, including different robustness penalties \cite{liu2021just,harmonization,krueger2021out,cha2021swad,mahajan2021domain,izmailov2022feature,rame2022fishr}, meta-learning \cite{zhang2020adaptive,collins2020task,aml,wang2021meta,jiang2023invariant}, unsupervised domain adaptation \cite{ganin2015unsupervised,long2016unsupervised,xu2021cdtrans}, dropout \cite{labonte2022dropout}, flat minima \cite{cha2021swad,flatminima}, weight averaging \cite{rame2022diverse,wortsman2022model,lawa}, (counterfactual) data augmentation \cite{survey, gowal2021improving, yao_improving_2022, yin2023ttida}, fine-tuning of only specific layers \cite{kirichenko2022last,lee_surgical_2023}, diversity \cite{teney2022evading, rame2022diverse}, etc.

\section{Conclusion}
\label{sec:conclusion}

We present \benchmark, an image classification benchmark with two types of spurious correlations, one-to-one (O2O) and many-to-many (M2M). We carefully design six dataset desiderata and instantiate them by leveraging recent advances in text-to-image and image  captioning models. Next, we  conduct experiments, and our findings indicate that even state-of-the-art group robustness techniques are insufficient in handling \benchmark, particularly in scenarios with Hard-splits where accuracy is below $70\%$. Our analysis of model errors revealed a dependence on irrelevant backgrounds, thus underscoring the difficulty of our dataset and highlighting the need for further investigations in this area.

\bibliographystyle{icml}
\bibliography{references}


\newpage

\appendix
\section{Effect of Model Architectures} \label{sec:architectures}

\begin{table*}[h!]
        \centering
        \resizebox{\textwidth}{!}{
        \begin{tabular}{c|ccc|ccc|c}
        \toprule
        \bf \multirow{2}{*}{Method} & \multicolumn{3}{c|}{\bf One-To-One SC} &  \multicolumn{3}{c|}{\bf Many-To-Many SC}  & \bf \multirow{2}{*}{Average}\\
        \cmidrule{2-7}
        & \bf Easy & \bf Medium & \bf Hard & \bf Easy & \bf Medium & \bf Hard &   \\
        \midrule
        ViT-B/16 (ERM) & $69.06\%_{\pm 0.05}$ & $68.58\%_{\pm 0.01}$ & $59.38\%_{\pm 0.02}$ & $71.46\%_{\pm 0.02}$ & $43.16\%_{\pm 0.03}$ & $44.99\%_{\pm 0.02}$ & $59.44\%$\\
        ViT-B/16 (MMD-AAE) & $71.29\%_{\pm 0.06}$ & $65.40\%_{\pm 0.01}$ & $60.83\%_{\pm 0.06}$  & $75.72\%_{\pm 0.01}$ & $46.45\%_{\pm 0.02}$ & $42.46\%_{\pm 0.03}$ & $60.36\%$\\
        \bottomrule
        \end{tabular}}

     \caption{\textbf{Effect of Different Architectures} on \benchmark-\{O2O,M2M\}-\{Easy, Medium, Difficult\}}
     \label{tab:results_new_archs}
\end{table*}

We experiment with the ViT-B/16 \cite{vit}, following \cite{izmailov2022feature,mehta2022you}. 
Based on \Cref{tab:results_new_archs}, we make the following observations:
The ViT backbone architecture worsens the performance for both MMD-AAE and ERM, underperforming the ResNet18 and the larger ResNet50. The best results were obtained by the CORAL algorithm applied to a ResNet50 which performs 17.1\% better than when than the best ViT. In the debate on whether ViTs \cite{vit} are generally more robust to SCs \cite{vitspurious} than CNNs or not \cite{izmailov2022feature,mehta2022you}, our results side with the latter. We observe that a ViT-B/16 pretrained on ImageNet22k had worse test accuracies than the ResNet architectures. 

\section{Appendix}

Include extra information in the appendix. This section will often be part of the supplemental material. Please see the call on the NeurIPS website for links to additional guides on dataset publication.

\begin{enumerate}

\item Submission introducing new datasets must include the following in the supplementary materials:
\begin{enumerate}
  \item Dataset documentation and intended uses. Recommended documentation frameworks include datasheets for datasets, dataset nutrition labels, data statements for NLP, and accountability frameworks.
  \item URL to website/platform where the dataset/benchmark can be viewed and downloaded by the reviewers.
  \item Author statement that they bear all responsibility in case of violation of rights, etc., and confirmation of the data license.
  \item Hosting, licensing, and maintenance plan. The choice of hosting platform is yours, as long as you ensure access to the data (possibly through a curated interface) and will provide the necessary maintenance.
\end{enumerate}

\item To ensure accessibility, the supplementary materials for datasets must include the following:
\begin{enumerate}
  \item Links to access the dataset and its metadata. This can be hidden upon submission if the dataset is not yet publicly available but must be added in the camera-ready version. In select cases, e.g when the data can only be released at a later date, this can be added afterward. Simulation environments should link to (open source) code repositories.
  \item The dataset itself should ideally use an open and widely used data format. Provide a detailed explanation on how the dataset can be read. For simulation environments, use existing frameworks or explain how they can be used.
  \item Long-term preservation: It must be clear that the dataset will be available for a long time, either by uploading to a data repository or by explaining how the authors themselves will ensure this.
  \item Explicit license: Authors must choose a license, ideally a CC license for datasets, or an open source license for code (e.g. RL environments).
  \item Add structured metadata to a dataset's meta-data page using Web standards (like schema.org and DCAT): This allows it to be discovered and organized by anyone. If you use an existing data repository, this is often done automatically.
  \item Highly recommended: a persistent dereferenceable identifier (e.g. a DOI minted by a data repository or a prefix on identifiers.org) for datasets, or a code repository (e.g. GitHub, GitLab,...) for code. If this is not possible or useful, please explain why.
\end{enumerate}

\item For benchmarks, the supplementary materials must ensure that all results are easily reproducible. Where possible, use a reproducibility framework such as the ML reproducibility checklist, or otherwise guarantee that all results can be easily reproduced, i.e. all necessary datasets, code, and evaluation procedures must be accessible and documented.

\item For papers introducing best practices in creating or curating datasets and benchmarks, the above supplementary materials are not required.
\end{enumerate}

\section*{Checklist}

\begin{enumerate}

\item For all authors...
\begin{enumerate}
  \item Do the main claims made in the abstract and introduction accurately reflect the paper's contributions and scope?
    \answerYes{}
  \item Did you describe the limitations of your work?
    \answerYes{} See \Cref{sec:conclusion}
  \item Did you discuss any potential negative societal impacts of your work?
    \answerNA{}
  \item Have you read the ethics review guidelines and ensured that your paper conforms to them?
    \answerYes{}
\end{enumerate}

\item If you are including theoretical results...
\begin{enumerate}
  \item Did you state the full set of assumptions of all theoretical results?
    \answerNA{}
	\item Did you include complete proofs of all theoretical results?
    \answerNA{}
\end{enumerate}

\item If you ran experiments (e.g. for benchmarks)...
\begin{enumerate}
  \item Did you include the code, data, and instructions needed to reproduce the main experimental results (either in the supplemental material or as a URL)?
    \answerYes{} See https://github.com/aengusl/spawrious
  \item Did you specify all the training details (e.g., data splits, hyperparameters, how they were chosen)?
    \answerYes{} see \Cref{sec:experiments}
	\item Did you report error bars (e.g., with respect to the random seed after running experiments multiple times)?
    \answerYes{} see \Cref{tab:results}
	\item Did you include the total amount of compute and the type of resources used (e.g., type of GPUs, internal cluster, or cloud provider)?
    \answerNo{} 
\end{enumerate}

\item If you are using existing assets (e.g., code, data, models) or curating/releasing new assets...
\begin{enumerate}
  \item If your work uses existing assets, did you cite the creators?
    \answerYes{} We cite the pretrained models and architectures we use
  \item Did you mention the license of the assets?
    \answerNo{} License is in the github repository https://github.com/aengusl/spawrious
  \item Did you include any new assets either in the supplemental material or as a URL?
    \answerNo{}
  \item Did you discuss whether and how consent was obtained from people whose data you're using/curating?
    \answerNA{}
  \item Did you discuss whether the data you are using/curating contains personally identifiable information or offensive content?
    \answerNA{}
\end{enumerate}

\item If you used crowdsourcing or conducted research with human subjects...
\begin{enumerate}
  \item Did you include the full text of instructions given to participants and screenshots, if applicable?
    \answerNA{}
  \item Did you describe any potential participant risks, with links to Institutional Review Board (IRB) approvals, if applicable?
    \answerNA{}
  \item Did you include the estimated hourly wage paid to participants and the total amount spent on participant compensation?
    \answerNA{}
\end{enumerate}

\end{enumerate}

\end{document}